\documentclass[akbc,twoside,11pt,lettersize]{article}
\usepackage{akbc}
\akbcheading
\ShortHeadings{Knowledge Graph Simple Question Answering for~Unseen~Domains}{Sidiropoulos, Voskarides,  \& Kanoulas}
\finalcopy 

\usepackage{booktabs}
\usepackage{wrapfig}

\usepackage{amsmath}
\usepackage{amssymb}

\usepackage{tikz}
\usetikzlibrary{shapes,arrows}

\usepackage{multirow}
\usepackage{tabularx}

\newcommand{\nv}[1]{\textcolor{orange}{[#1]}}
\newcommand{\gs}[1]{\textcolor{green}{[#1]}}
\newcommand{\todo}[1]{\textcolor{black}{#1}}

\usepackage[font=small,labelfont=bf]{caption}
\usepackage{subcaption} 
\begin{document}

\title{Knowledge Graph Simple Question Answering for~Unseen~Domains}

\author{\name Georgios Sidiropoulos 
        \email g.sidiropoulos@uva.nl \\
        \name Nikos Voskarides 
        \email nickvosk@gmail.com \\
        \name Evangelos Kanoulas 
        \email e.kanoulas@uva.nl \\
       \addr University of Amsterdam, The Netherlands
       }
\maketitle

\begin{abstract}

Knowledge graph \todo{simple} question answering (\todo{KGSQA}), in its standard form, does not take into account that human-curated \todo{question answering} training data only cover a small subset of the relations that exist in a Knowledge Graph (KG), or even worse, that new domains covering unseen and rather different to existing domains relations are added to the KG. In this work, we study KGSQA in a previously unstudied setting where new, unseen domains are added during test time.
In this setting, question-answer pairs of the new domain do not appear during training, thus making the task more challenging.
We propose a data-centric domain adaptation framework that consists of a KGSQA system that is applicable to new domains, and a sequence to sequence question generation method that automatically generates question-answer pairs for the new domain.
Since the effectiveness of question generation for KGSQA can be restricted by the limited lexical variety of the generated questions, we use distant supervision to extract a set of keywords that express each relation of the unseen domain and incorporate those in the question generation method.
Experimental results demonstrate that our framework significantly improves over  zero-shot baselines and is robust across domains.

\end{abstract}

\maketitle

\section{Introduction}
%
Large-scale structured Knowledge Graphs (KGs) such as Freebase~\cite{bollacker2008freebase} and Wikidata~\cite{pellissier2016freebase} store real-world facts in the form of subject--relation--object triples.
KGs are being increasingly used in a variety of tasks that aim to improve user experience~\cite{bota2016playing}. One of the most prominent tasks is Knowledge Graph Simple Question Answering (KGSQA),
which aims to answer natural language questions by retrieving KG facts~\cite{yih2015semantic}. 
In practice, many questions can be interpreted by a single fact in the KG. This has motivated the KGSQA task~\cite{bordes2015large,Mohammed2018StrongBF,Petrochuk2018SimpleQuestionsNS}, which is the focus of this paper. In KGSQA, given a \textit{simple} question, e.g. \todo{``who directed the godfather?'', the system should interpret the question and arrive at a single KG fact that answers it: (The Godfather (film), film.film.directed\_by, Francis Ford Coppola)}.

KGSQA systems are trained on manually annotated datasets that consist of question-fact pairs. In practice, the applicability of such systems in the real-world is limited by two factors: (i) modern KGs store millions of facts that cover thousands of different relations, but KGSQA training datasets can only cover a small subset of the existing relations in the KG ~\cite{elsahar2018zero}, and (ii) KGs are dynamic, i.e. they are updated with new domains that cover new relations~\cite{pellissier2016freebase}.
Solving (i) and (ii) by exhaustively gathering question-fact pair annotations would be prohibitively laborious, thereby we need to rely on automatic methods. 

Motivated by the above, in this work, we study the KGSQA task in a setting where we are interested in answering questions about a new, unseen domain that covers relations, for which we \textit{have} instances in the KG, but we \textit{have not seen} any question-fact pair during training.
We model this as a domain adaptation task~\cite{mansour2009domain, pan2009survey} and propose a data-centric domain adaptation framework to address it.
Data-centric domain adaptation approaches focus on transforming or augmenting the training data, instead of designing specialized architectures and training objectives as model-centric domain adaptation approaches do~\cite{chu2018survey}.
Our framework consists of: (a) a KGSQA system which can handle the unseen domain, and (b) a novel method that generates training data for the unseen test domain.
The KGSQA system we introduce performs mention detection, entity candidate generation and relation prediction on the question, and finally selects the fact that answers the question from the KG.
To improve relation prediction on questions that cover relations of the  unseen domain, we automatically generate synthetic questions from KG facts of the unseen domain (i.e.\ knowledge graph question generation -- QG).
The resulting synthetic question-fact pairs are used to train the KGSQA system for the unseen domain.
We find that the effectiveness of QG for KGSQA can be restricted not only by the quality of the generated questions, but also by the lexical variety of the questions.
This is because users ask questions underlying the same relation using different lexicalizations (e.g. ``who is the author of X'', ``who wrote X'').
To address this, we use distant supervision to extract a set of keywords for each relation of the unseen domain and incorporate those in the question generation method.

Our main contributions are the following: (i) we introduce a new setting for the KGSQA task, over new, previously unseen domains,
(ii) we propose a data-centric domain adaptation framework for KGSQA that is applicable to unseen domains, and (iii) we use distant supervision to extract a set of keywords that express each relation of the unseen domain and incorporate them in QG to generate questions with a larger variety of relation lexicalizations.
We experimentally evaluate our proposed method on a large-scale KGSQA dataset that we adjust for this task and show that our proposed method consistently improves performance over zero-shot baselines and is robust across domains.\footnote{Our code is available at \url{https://github.com/GSidiropoulos/kgsqa_for_unseen_domains}.}

\section{Problem Statement}
\label{sec:prob-statement}
Let $E$ denote the set of entities and $R$ the set of relations.
A KG $K$ is a set of facts $(e_s, r, e_o)$, where $e_s, e_o \in E$  are the subject and object entities respectively, and $ r \in R$ is the relation between them. 
Each relation $r$ has a unique textual label $r_l$ and falls under a single domain $\mathcal{D}$.
For instance, music.album.release\textunderscore type and music.artist.genre fall under the Music domain.
\textit{Simple} questions mention a single entity and express a single relation.
For instance, the question \todo{``who directed the godfather?'' mentions the entity ``The Godfather'' and expresses the relation  film.film.directed\_by}.
Given a \textit{simple} question $q$ that consists of a sequence of \todo{tokens} $t_1,t_2,\dots, t_T$,  the KGSQA task is to retrieve a fact $(\hat{e}_s, \hat{r}, \hat{e}_o)$, where $(\hat{e}_s, \hat{r})$ accurately interprets $q$ (i.e., $\hat{e}_s$ is mentioned in $q$ and $\hat{r}$ is expressed in $q$) while $\hat{e}_o$ provides the answer to $q$. 
In our setting, we aim to build a KGSQA system that can perform well on a previously unseen domain. A domain is ``unseen'' when facts that cover relations of that domain do exist in $K$, but gold-standard question-fact pairs of that domain do \textit{not} appear in the training data.
This setting is an instance of domain adaptation, where a model is trained on data $\mathcal{S}$, which is drawn according a source distribution, and tested on data $\mathcal{T}$ coming from a different target distribution. 
Domain adaptation over KG domains is more challenging compared to domain adaptation over single KG relations~\cite{Yu2017ImprovedNR, WuHWZZYC19}, because it is less likely for relations with similar lexicalizations to appear in the training set.

\section{KGSQA system}
\label{sec:KBQA-system}
In this section, we detail our KGSQA system.
In Section~\ref{sec:QG}, we will describe how we generate synthetic training data to make this system applicable to unseen domains.
Following current state-of-the-art on KGSQA~\cite{Petrochuk2018SimpleQuestionsNS}, we split the task into four sub-tasks, namely, \textit{entity mention detection} (MD), \textit{entity candidate generation} (CG), \textit{relation prediction} (RP), and \textit{answer selection} (AS).
The skeleton of our KGSQA system generally follows previous work, and we modify the MD and RP architectures.
\label{sec:KBQA}
\paragraph{Mention Detection (MD)}
Given the question $q$, MD outputs a single entity mention $m$ in $q$, where $m$ is a sub-sequence of tokens in $q$.
We model this problem as sequence tagging, where given a sequence of tokens, the task is to assign an output class for each token~\cite{Huang2015BidirectionalLM,Lample2016NeuralAF}. 
In our case, the output classes are entity (E) and context (C).
For instance, the correct output for the question \todo{``who directed the godfather?" is ``[C C E E]''}.
We use a BiLSTM with residual connections (R-BiLSTM) \cite{HeZRS16}, since it outperformed  vanilla RNN, BiRNN, and a CRF on top of a BiRNN \cite{Petrochuk2018SimpleQuestionsNS} in preliminary experiments.
\paragraph{Candidate Generation (CG)}
\label{sec:cg}
Given the mention $m$ extracted from the previous step, CG maps $m$ to a set of candidate entities $C_S \subset E$. 
For instance, \todo{CG maps the mention ``the godfather'' to the entities $\{$ The Godfather(film), The Godfather(book)\dots $\}$}.
The CG method we use was proposed in \citet{Tre2016NoNT}.
Briefly, the method pre-builds an inverted index $I$ from n-grams of mentions to entities, and it looks-up the n-grams of $m$ in $I$ to obtain $C_S$.

\paragraph{Relation Prediction (RP)}
\label{sec:rp}
Given the question $q$ and the set of entities $C_s$ extracted in the previous step, RP outputs a single relation $\hat{r} \in R$ that is expressed in $q$.
Previous work models RP as a large-scale multi-label classification task where the set of output classes is fixed \cite{Petrochuk2018SimpleQuestionsNS}.
In our domain adaptation scenario, however, we want to be able to predict relations that we have not seen during training. 
Therefore, we model RP as a relation ranking task, as in~\cite{Yu2017ImprovedNR}, and use the textual label $r_l$ to represent the relation $r$ (instead of using a categorical variable). This way we can in principle represent any relation $r \in R$ during inference time.
Below we describe the architecture we use for RP and how we perform training and inference.

Our architecture is a simpler version of \cite{Yu2017ImprovedNR}, where they model a relation both as a sequence and a categorical variable, and they use more complex sequence encoders. 
First we describe how we encode the question $q$ and the relation $r$.
In order to generalize beyond specific entity names, we first replace the previously detected entity mention $m$ in $q$ with a placeholder token; e.g ``who directed SBJ''.
We then map each term to its embedding and feed the word embeddings to an LSTM; embeddings are initialized with pretrained word2vec
embeddings \cite{mikolov2013distributed}. The final hidden state of the LSTM $\boldsymbol{\gamma}^{(q)}$ is used as the encoding of the question.
In order to represent $r$, we use its label $r_l$ (e.g. film.film.directed\_by). Similarly with the question encoding, we encode $r_l$ with an LSTM to obtain $\boldsymbol{\gamma}^{(r)}$. However, since questions and relations significantly differ both grammatically and syntactically, the two LSTM encoders do not share any parameters. The ranking function $f$ is calculated as $f(q,r)=\cos{(\boldsymbol{\gamma}^{(q)},\boldsymbol{\gamma}^{(r)})}$,
where $\cos(\cdot)$ is the cosine similarity.
We train $f$ using standard pairwise learning to rank.
The loss is defined as follows:
\begin{align}
	L(\theta) = \sum_{r} \sum_{r' \in R'}   \max(0, \mu - f(q, r) + f(q, r')),
\end{align} 
where $\theta$ are the parameters of the model, $\mu$ is a hyperparameter, and $R'$ is the set of sampled negative relations for a question $q$.
We design a specialized \textit{negative sampling} method to select $R'$. With probability $P_{R}^{-}$ we uniformly draw a sample from $R^{-}=\{r'|r' \in R \land r' \neq r \}$; the set of all available relations except the positive relation $r$. 
With probability $1-P_{R}^{-}$ we draw a random sample from $\hat{R}^{-}=\{r'|r' \in \mathcal{D}_{R}^{+} \land r' \neq r \}$; the set of relations that are in the same domain as the positive relation $r$.
This way, we expose the model to conditions it will encounter during inference.
At inference time, given a question $q$ and a set of relations we score all question-relation pairs ($q, r$) with $f$ and select the relation $\hat{r}$ with the highest score.
Unfortunately, computing a score with respect to all possible relations in $R$ leads to poor performance when there is no linguistic signal to disambiguate the choice.
In order to address this issue, we constrain the set the potential output relations  $R_c$ to be the union of the relations expressed in the facts where the entities in $C_S$ participate in~\cite{Petrochuk2018SimpleQuestionsNS}. Formally, we define the target relation classes to be $R_c =\{ r \in R | (e_s,r,e_o) \in K \wedge e_s \in C_S \}$. \todo{For example, given the question ``who directed the godfather'', the potential relations are $\{$ film.film.directed\_by, book.written\_work.author, $\dots \}$. Using the aforementioned constraint we can safely ignore relations like tv.tv\_series\_episode.director by taking into account that The Godfather does not appear in any tv-related facts.}
\paragraph{Answer Selection (AS)}
Given the set of entities $C_S$ obtained from CG, and the top ranked relation $\hat{r}$ obtained from RP, AS selects a single fact $(\hat{e}_s,\hat{r},\hat{e}_o)$, where $\hat{e}_o$ answers the question $q$. 
The set of candidate answers may contain more than one facts $(e'_s, \hat{r}, e'_o)$, where $\forall e'_s \in C_S$.
Since there is no explicit signal on which we can rely to disambiguate the choice of subject, all the potential answers are equally probable.
We therefore use a heuristic based on popularity, \todo{introduced by} \citet{Mohammed2018StrongBF}: we choose $\hat{e}_s$ to be the entity that appears the most in the facts in $K$ either as a subject or as an object.
Having $\hat{e}_s$ and $\hat{r}$ we can retrieve the fact $(\hat{e}_s,\hat{r},\hat{e}_o)$. 
\todo{For our running example (``who directed the godfather"), given film.film.directed\_by (from RP) and entities $\{$ The Godfather(film), The Godfather(book)\dots $\}$ (from CG) we can select the fact (The Godfather (film), film.film.directed\_by, Francis Ford Coppola).}

\section{KGSQA to unseen domains using question generation}
\label{sec:QG}

Even though all the components of the aforementioned KGSQA system were designed to work with unseen domains, preliminary experiments demonstrated that RP does not generalize well to questions originating from unseen domains.
This is expected since RP is a large-scale problem (thousands of relations), and it is very challenging to model less frequent or even unknown relations that are expressed with new lexicalizations.

We therefore focus on improving RP for questions originating from unseen domains.
Inspired from the recent success of data-centric domain adaptation in neural machine translation~\cite{chu2018survey}, we perform synthetic question generation from KG facts of the unseen domain to generate question-fact pairs for training the RP component (see Section~\ref{sec:rp}).\footnote{Note that \citet{DongMRL17} also performed QG for improving the overall KGSQA performance. However, their model is not applicable to our domain adaptation scenario since their model relies on modifying existing questions and all domains were predefined.}
In the remainder of this section we briefly describe the base question generation (QG) model we build upon and how we augment the model to more effectively use textual evidence and thus better generalize to relations of the unseen domain. 
\subsection{Base model for QG}
\label{sec:kgqg-model}

Given a fact $(e_s, r, e_o)$ from the target domain, QG aims to generate a synthetic question $\hat{q}$.
During training, only question-fact pairs from the known domains are used.
Our base model is the state-of-the-art encoder-decoder architecture for QG~\cite{elsahar2018zero}. 
It takes as input the fact $(e_s, r, e_o)$ alongside with a set of textual contexts $C=\{c_s, c_r, c_o\}$ on the fact.
Those textual contexts are obtained as follows: $c_s$ and $c_o$ are the types of entities $e_s$ and $e_o$ respectively, whereas $c_r$ is a lexicalization of the relation $r$ obtained by simple pattern mining on Wikipedia sentences that contain instances of $r$.
For instance, given the fact (The Queen Is Dead, music.album.genre, Alternative Rock), the textual contexts are: $c_s=\{\text{``album''}\}$, $c_r=\{\text{``album by''}\}$ and $c_o=\{\text{``genre''}\}$.

The encoder maps $e_s$, $r$ and $e_o$ to randomly initialized embeddings and concatenates those to encode the whole fact.
Also, it encodes the text in $c_s$, $c_r$ and $c_o$ separately using RNN encoders.
The decoder is a separate RNN that takes the representation of the fact and the RNN hidden states of the textual contexts to generate the output question $\hat{q}$.
It relies on two attention modules: one over the encoded fact and one over the encoded textual contexts.
The decoder generates tokens not only from the output vocabulary but also from the input (using a copy mechanism) to deal with unseen input tokens. 
\subsection{Using Richer Textual Contexts for QG}
\label{sec:kgqg-textual-contexts}
The role of the textual contexts $C$ in the aforementioned base model is critical, since it enables the model to provide new words/phrases that would have been unknown to the model otherwise~\cite{elsahar2018zero}.
Even though the base model generally generates high quality questions, in our task (KGSQA), we aim to generate a larger range of lexicalizations for a single relation during training in order to generalize better at test time.
This is because users with the same intent may phrase their questions using different lexicalizations (e.g. ``who is the author of X'', ``who wrote X'').
Thus, in this section we focus on how to provide the model with a diverse set of lexicalizations for a relation $r$ instead of a single one as in the base model, in order to be able to generate a more diverse set of questions in terms of relation lexicalizations.
More precisely, given a relation $r$, we extract $k$ keywords that will constitute the relation's textual context $c_r$.
To this end, we first extract a set of candidate sentences $S_r$ that express a specific relation $r$ between different pairs of entities. Second, we extract keywords from the set $S_r$, rank them and select the top-$k$ keywords that constitute the set $c_r$. We detail each of these steps below.

\paragraph{Extracting sentences}
\label{sec:extracting-sentences}
Given a set of facts $F_r$ of relation $r$ between different pairs of entities, we aim to extract a set of sentences $S_r$, where each sentence $s \in S_r$ expresses a single fact $(e_s,r,e_o)$ in $F_r$~\cite{voskarides-etal-2015-learning}.
For this, for each fact $(e_s,r,e_o)$ in $F_r$, we need to (a) extract a set of candidate sentences $S$ that might express $(e_s,r,e_o)$ and (b) select the sentence that best expresses the relation.
For (a), we collect the set of sentences $S$ using distant supervision, similarly to ~\cite{mintz2009distant}: $S$ consists of sentences that mention $e_o$ in the Wikipedia article of $e_s$ and sentences that mention $e_s$ in the Wikipedia article of $e_o$.
For (b), we score each sentence $s \in S$ w.r.t. the label $r_l$ of the relation $r$ using the cosine similarity $cos(e(s), e(r_l))$, 
where $cos(\cdot)$ is the cosine similarity and $e(x)$ is calculated as $e(x)=(1/|x|)\sum_{t \in x} w_t$,
where $w_t$ is the embedding of word $t$.
Finally, we take the sentence $s'$ with the highest score and add it to the set $S_r$.
\paragraph{Extracting keywords}
\label{par:keywords}
After extracting the set of sentences $S_r$, we aim to extract the set of keywords $c_r$.
For this, we treat $S_r$ as a single document and score each word $t$ that appears in $S_r$ using tf-idf, $\text{score}(t)= \text{tf}(t,S_r)\cdot \text{idf} (t,S_R)$,
where $S_R$ is the union of all $S_{r'}, r' \in R$.
The top-$k$ scoring words constitute the set of keywords $c_r$. 
Table \ref{tab:keywords} depicts example keywords generated by the procedure described above.

\paragraph{}
The keyword extraction approach described above is conceptually simple yet we later show that it significantly improves upon the base model when applied to KGSQA.
\begin{table*}[t]
\caption{Examples of relation textual contexts extracted by our keyword extraction approach.}
\small
\begin{tabularx}{\linewidth}{p{6cm} p{9cm}}
\toprule
Relation                                   & Textual Context                                                                    \\ \midrule
music.artist.label                       & records, artists, album, released, label, signed, band\\
film.film.directed\_by                  & film, director, directed, films, short, directing, producer\\
people.deceased\_person.place\_of\_death & died, death, deaths, born, age, people, male, actors\\
\bottomrule
\end{tabularx}
\label{tab:keywords}
\vspace{-4mm}

\end{table*}

\section{Experimental Setup}
\label{sec:exp-setup}
In this section, we discuss how we design the experiments to answer the following research questions:
\textbf{RQ1}) How does our method for generating synthetic training data for the unseen domain perform on RP compared to a set of baseline methods?
\textbf{RQ2}) How does our full method perform on KGSQA for unseen domains compared to state-of-the-art zero-shot data-centric methods? 
\textbf{RQ3}) How does our data-centric domain adaptation method compare to a state-of-the-art model-centric method on RP? 
\paragraph{Dataset}
\label{sec:dataset}
In our experiments we use the SimpleQuestions dataset, which is an established benchmark for studying KGSQA~\cite{bordes2015large}.
The dataset consists of 108,442 questions written in natural language by human annotators, paired with the ground truth fact that answers the question. The ground truth facts originate from Freebase~\cite{bollacker2008freebase}.
The dataset covers 89,066 unique entities, 1,837 unique relations and 82 unique domains.
In our setup, we leave one domain out to simulate a new, previously unseen domain, and train on the rest. 
We choose six challenging domains as target domains: Film, Book, Location, Astronomy, Education and Fictional Universe; the first three are among the largest domains and the last three are medium-sized.
The aforementioned domains are challenging because they have very low overlap in terms of relation lexicalization w.r.t. the rest of the domains used as source domains.
The training data consists of the question-fact pairs that  appear in the source domains, augmented with synthetically generated data of the target/unseen domain.
In practice, we replace all questions from the target domain that initially appear in the full training set with their corresponding synthetically generated questions.\footnote{One may hypothesize that since entities can appear in multiple domains (e.g. actors who are also singers), question generation becomes an unrealistically simple task. However, this is not the case because in our dataset, the entity overlap between seen and unseen domains is \emph{only} 4.6\%.}
As a source of text documents for the textual context collection (see Section~\ref{sec:kgqg-textual-contexts}), we use Wikipedia articles augmented with dense entity links provided by DAWT~\cite{Spasojevic:2017:DDA:3041021.3053367}.

\paragraph{Baselines}
To answer \textbf{RQ1}, we keep the KGSQA system unchanged and alternate the way of generating synthetic questions.
We compare the RP performance on the unseen domain given the following ways of generating synthetic data of the unseen domain: (i) No synthetic data, (ii) Wiki-raw-sentences: uses the raw Wikipedia sentence that expresses the ground truth fact that answers the question (automatically extracted using the procedure in Section~\ref{sec:extracting-sentences}), and (iii) the state-of-the-art QG method proposed in \citet{elsahar2018zero}.
To answer \textbf{RQ2}, we replace our RP component with two state-of-the-art RP models:
(i) \citet{Petrochuk2018SimpleQuestionsNS}, which uses a BiLSTM to classify relations, and (ii) \citet{Yu2017ImprovedNR}, a zero-shot RP model that uses a HR-BiLSTM and is specifically designed to deal with unseen or less frequently seen relations.
To answer \textbf{RQ3}, we compare the performance of our data-centric model on RP against a state-of-the-art model-centric zero-shot approach~\cite{WuHWZZYC19}: it the HR-BiLSTM proposed by \citet{Yu2017ImprovedNR} with an adversarial adapter combined and a reconstruction loss. The adapter uses embeddings trained on Freebase and Wikipedia by JointNRE \cite{Han0S18} and learns representations for both seen and unseen relations.
\paragraph{Evaluation metrics}
\label{sec:eval-metrics}
We run the experiments three times and report the median (only marginal and not significant differences were found among different runs)~\cite{Mohammed2018StrongBF}. %
In contrast to the classic KGSQA where the task is to retrieve a single entity, it is standard practice when using the SimpleQuestions dataset to treat the problem as question interpretation~\cite{Petrochuk2018SimpleQuestionsNS}. 
More specifically, the objective is to rewrite the natural language question in the form of subject-relation pair.
We evaluate our overall approach in terms of top-1 accuracy, i.e.\ whether the retrieved subject-relation pair matches the ground truth. 
We measure accuracy both at a macro- (domains) and at a micro-level (samples).
Statistical significance is determined using a paired two-sided t-test.

\paragraph{Parameter configurations}
\label{appendix_parameters}
We initialize word embeddings with pretrained Google News 300-dimensional embeddings \cite{mikolov2013distributed}. 
We use the Adam optimizer \cite{kingma2014adam}. 
Our MD model consists of 2 hidden layers, 600 hidden units, 0.4 dropout rate, frozen embeddings, and learning rate of $10^{-3}$; 50 training epochs.
For the RP model, we use 1 layer encoder for both questions and relations that consists of $400$ hidden units, with a frozen embedding layer, and a learning rate of $10^{-3}$; trained for 10 epochs. 
During training, we sample 10 negative questions per question using the procedure described in Section~\ref{sec:rp}.
We use a batch size of 300 and 200 for MD and RP respectively.
For the QG model~\cite{elsahar2018zero} and the model-centric RP~\cite{WuHWZZYC19} model we compare against, we use the hyperparameters as presented in their work. 
Note that for both our method and the baselines, the hyperparameters were tuned on the initial split of the SimpleQuestions dataset. We keep the parameters fixed for both our method and the baselines for all source-target domain setups.
We set the number of keywords for each relation $k=10$ (Section~\ref{sec:kgqg-textual-contexts}).

\section{Results and Discussion}
\label{sec:results}
In this section we present and discuss our experimental results. All models under comparison have all their components fixed, except RP. Therefore, any improvement observed, is due to RP.
\begin{table}[t]
\centering
\caption{Relation Prediction accuracy w.r.t. different ways of generating synthetic training data for the unseen domain. $\blacktriangle$ indicates a significant increase in performance compared to the top performing baseline  ($p <0.01$).}
\small
\begin{tabular}{@{}lll@{}}
\toprule
\begin{tabular}[c]{@{}l@{}}Synthetic\\training data\end{tabular} & \begin{tabular}[c]{@{}l@{}}Macro-avg.\\ Accuracy (\%)\end{tabular} & \begin{tabular}[c]{@{}l@{}}Micro-avg.\\ Accuracy (\%)\end{tabular} \\ \midrule
- & 30.21 & 29.06 \\
Wiki-raw-sentences & 37.89 & 36.51 \\
QG ~\cite{elsahar2018zero} & 67.52 & 69.78 \\
QG (Ours) & \bf{69.86}$^\blacktriangle$ & \bf{70.95} $^\blacktriangle$ \\
\bottomrule
\end{tabular}
\label{tab:RP_baselines}
 \vspace{-4mm} 

\end{table}

\begin{figure}[t]
    \centering
    \includegraphics[scale=0.55]{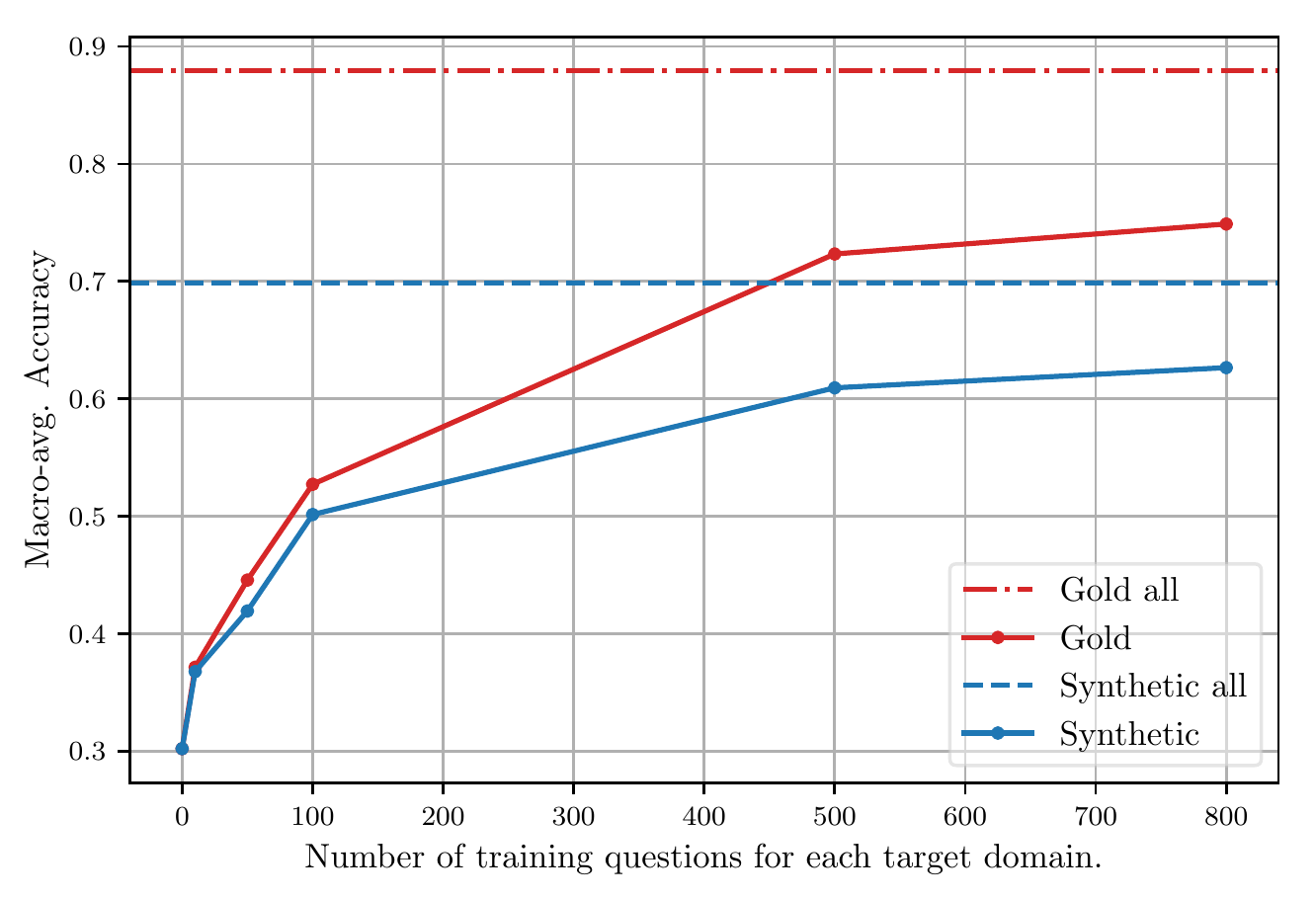}
    \caption{\todo{RP macro-accuracy when varying the number of target domain questions used to augment the training set. Gold refers to the gold standard questions and synthetic refers to the automatically generated questions. Gold all (Synthetic all) refers to the full set of training gold (synthetic) questions. The training set size of the smallest domain is 800, thus we report performance up to that point.}}
    \label{fig:rp-variation}
         \vspace{-4mm}
\end{figure}

\paragraph{Effect of synthetic data on RP (RQ1)}
Here we compare the RP performance of our method for generating synthetic training data with a set of baselines.
For this experiment, the RP component of the KGSQA system remains unchanged and we only alter the data it is trained with.
Table~\ref{tab:RP_baselines} shows the results.
We observe that our QG method is the best performing one.
It significantly outperforms the baseline QG method, which confirms that our method for generating rich textual contexts for relations (Section~\ref{sec:kgqg-textual-contexts}) is beneficial for KGSQA.
As expected, Wiki-raw-sentences performs better than when not using training data from the target domain at all but performs much worse than the QG methods.
This is expected since Wikipedia sentences are very different both syntactically and grammatically from the real questions that the KGSQA system encounters during test time.
\todo{Next, we investigate how RP performance varies depending on the number of the target domain questions used to augment the training set. Figure~\ref{fig:rp-variation} shows the results. First, we observe that, in the low data regime (less than 100 questions), the gap in performance between training with gold or synthetic questions is small. This is encouraging for applying our framework on domains in the long tail. As the number of questions increases, the performance for both gold and synthetic increase, however the gap between them increases, which is expected.}

\begin{figure}[t]
    \centering
    \includegraphics[scale=0.45]{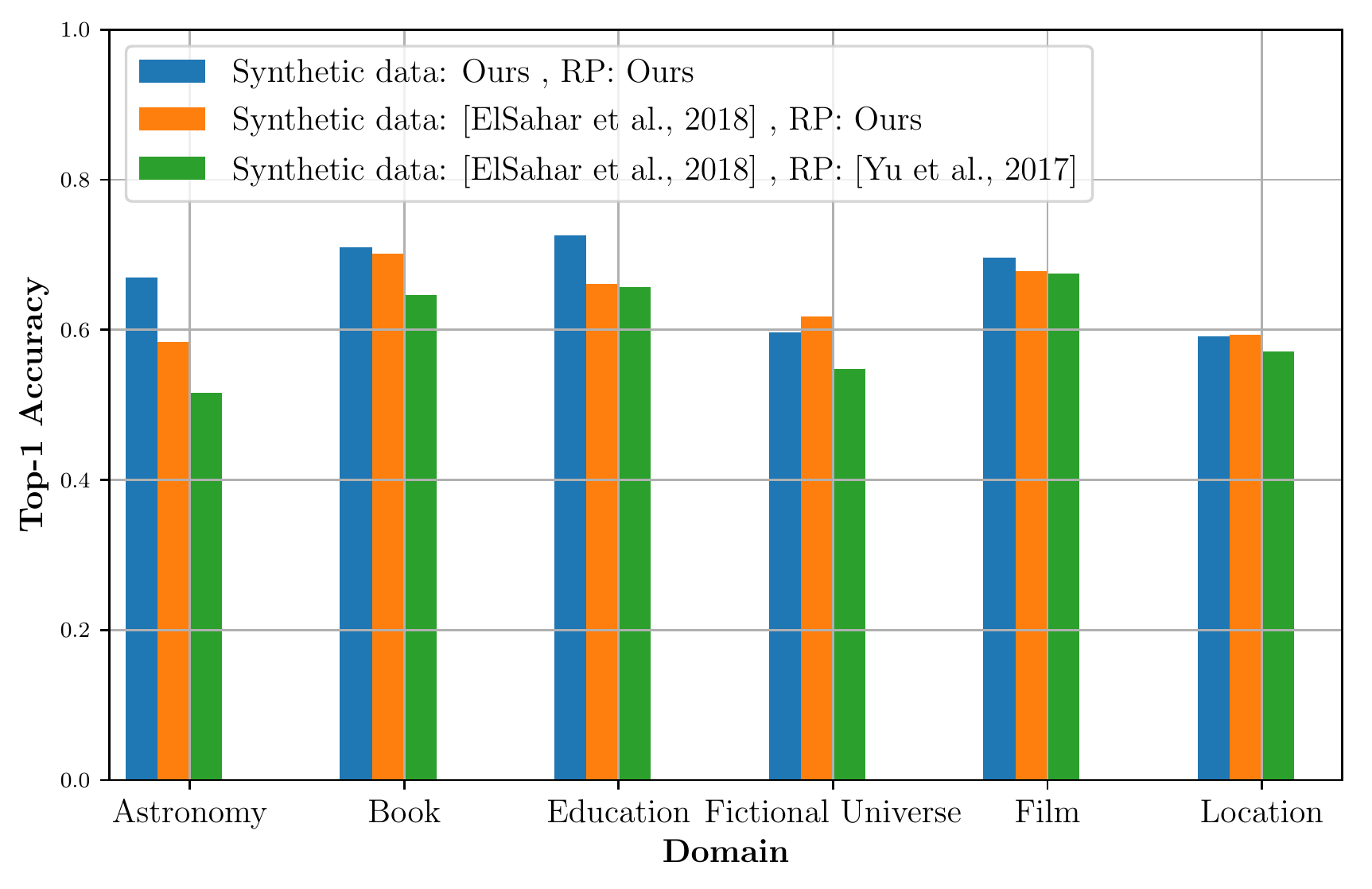}
        \caption{\todo{End-to-end accuracy on the KGSQA task per domain.}}
        \label{fig:KGSQA-accuracy-domain}

     \vspace{-4mm}

\end{figure}

\begin{table}[t]
\centering
\caption{End-to-end accuracy on the KGSQA task. $^\blacktriangle$ indicates a significant increase in performance compared to the top performing baseline  ($p <0.01$). $\dagger$ is for \citet{elsahar2018zero}, $\ddagger$ for ~\citet{Petrochuk2018SimpleQuestionsNS}, and $\star$ for ~\citet{Yu2017ImprovedNR}.
}
\small
\begin{tabular}{@{}llll@{}}
\toprule
\begin{tabular}[c]{@{}l@{}}Synthetic\\training data\end{tabular} & \begin{tabular}[c]{@{}l@{}}Relation\\ Prediction (RP)\end{tabular} & \begin{tabular}[c]{@{}l@{}}Macro-avg.\\ Accuracy (\%)\end{tabular} & \begin{tabular}[c]{@{}l@{}}Micro-avg.\\ Accuracy (\%)\end{tabular} \\ 
\midrule
\multirow{3}{*}{QG $\dagger$}
 & BiLSTM $\ddagger$ & 55.49 & 55.11 \\
 & HR-BiLSTM $\star$ & 60.20 & 62.77 \\
 & Ours &63.90& 65.18\\ 
\midrule
QG (Ours) & Ours& \textbf{66.49}$^\blacktriangle$&\textbf{66.64}$^\blacktriangle$\\ 
\midrule
Gold Questions & Ours & 84.56 & 82.87\\ 
\bottomrule
\end{tabular}
\label{tab:end2end}
     \vspace{-4mm}
\end{table}

\paragraph{Overall KGSQA performance for data-centric methods (RQ2)}
Next, we compare our full framework to variations that use state-of-the-art RP models.
Table~\ref{tab:end2end} shows the results.
We observe that our full method (second to last row) improves over all the baselines and significantly outperforms the best performing baseline.
As expected, we see that even though our full method holds strong generalization ability for unseen domains, there is a gap in the performance when using the automatically generated synthetic questions (second to last row) or the human generated questions (last row).
This gap suggests that there is room for improvement for QG.
Next, we test the systems under comparison in terms of generalization ability across domains. 
Figure~\ref{fig:KGSQA-accuracy-domain} shows the results.
First, we observe that our method achieves an accuracy of at least 60\% for all domains which shows that it is robust across domains. Also, it outperforms the baselines in all but one domain.
In order to gain further insights, we sampled success and failure cases from the test set.
We found that the errors in the failure cases generally originate from the fact that the model relies on lexicalizations that are frequent in the seen domains.
\todo{We show such cases in Table \ref{tab:quest_ex}}.
Furthermore, our analysis showed that one way of improving QG is to improve  keyword extraction by collecting a larger set of relevant sentences that express a single relation, possibly by looking into other sources of text (e.g. news articles). \todo{In Table \ref{tab:bleu}, the automatic evaluation results for the synthetic questions generated by our QG model against those generated by \cite{elsahar2018zero} further strengthen our claim. As can be seen from the table, our model outperforms the baseline, indicating that using a larger set of relevant sentences for a single relation, can be beneficial to the generated questions.}

\begin{table*}[h]
\small
\caption{\todo{Examples of success cases (top 3 rows) and failure cases (bottom 3 rows) of our QG method.}}
\todo{
\begin{tabularx}{\linewidth}{p{2cm} p{6cm} p{6cm}}
\toprule
Unseen \\Domain & Gold Questions & Synthetic Questions \\
\midrule
Astronomy & what is something that carolyn shoemaker discovered & what is the astronomical objects discovered by carolyn shoemaker \\
Book & what's the subject of the cognitive brain & what is the subjects of the written work the cognitive brain \\
Location & which country is cumberland lake located in & where is cumberland lake located  \\ 
\midrule
Film & in what country did the film joy division take place & what country is joy division under \\
Book & who authored the book honor thyself & who was the director of the book honor thyself\\
Location & what was a historic atlantic city convention hall team & what event took place at historic atlantic city convention hall in 1943\\ 
\bottomrule
\end{tabularx}}
\label{tab:quest_ex}
 \vspace{-4mm}
\end{table*}

\begin{table}[t]
\small
\centering
\caption{\todo{Automatic evaluation of question generation w.r.t. BLEU.}}
\todo{
\begin{tabular}{@{}lllll@{}}
\toprule
     & BLEU-1 & BLEU-2 & BLEU-3 & BLEU-4 \\ \midrule
QG (\cite{elsahar2018zero})  & 44.04  & 28.63  & 16.50  & 9.13   \\
QG (Ours) & \bf 44.27  & \bf 29.55  & \bf 17.73  & \bf 10.16  \\ \bottomrule
\end{tabular}}
\label{tab:bleu}
     \vspace{-4mm}
\end{table}

\paragraph{Comparison to a model-centric method (RQ3)}
Here, we compare our data-centric method for domain adaptation to a state-of-the-art model-centric method on RP~\cite{WuHWZZYC19}.
In order to perform a fair comparison when testing for RP, we follow their setup (see Section 5.1. in ~\cite{WuHWZZYC19}) and for this particular experiment we assume that MD and CG produce the correct output.
Our method outperforms their method both on macro-accuracy ($75.54\%$ vs $75.02\%$), and micro-accuracy ($77.08\%$ vs $72.17\%$).
Note that we use randomly initialized embeddings whereas in their work they use JointNRE relation embeddings trained on Wikipedia and Freebase, which provides an advantage to their method.
Also note that their method (model-centric) is orthogonal to ours (data-centric) and therefore, an interesting future work direction would be to explore how to combine the two methods to further improve performance. 

\paragraph{Qualitative error analysis}
\todo{
In order to gain insights on how each part of the pipeline affects the final prediction, we perform an empirical error analysis. 
We sample 60 examples for which our system provided a wrong answer (10 for each target domain), and investigate what led to the wrong prediction.
Out of these examples, 43 mistakes were due to RP, 6 due to ED, 5 due to CG and 6 due to AS.
For RP, 22 were assigned to conceptually similar relations within the target domain and 15 to similar relations  outside the target domain, while the rest were assigned to a common relation outside the target domain.
For ED, 4 were due to predicting an extra token as part of the entity mention and 2 were due to missing a token from the entity mention.
For CG, 2 were because the gold entity was not part of the mention candidates and 1 was due to an error in the human annotation; the early termination proposed in \cite{Tre2016NoNT}, is responsible for the rest.
From this analysis, we confirm that RP remains the most challenging part of KGSQA. 
Within RP, what seems to be the challenge is that there are relations for which there is a high lexical similarity between the corresponding questions, but also between the relations per se.}

\paragraph{KGSQA performance on seen domains} Finally, even though the focus of this paper is to perform KGSQA on unseen domains and thus we do not aim to improve state-of-the-art on seen domains, we also test our KGSQA system on the standard split of the SimpleQuestions dataset.
Our model achieves a top-1 accuracy of $77.0\%$, which is ranked third among the state-of-the-art methods while having a simpler method than the two top-performing ones (\cite{Petrochuk2018SimpleQuestionsNS}), \cite{gupta2018retrieve}). We provide a thorough comparison w.r.t. the state-of-the-art on seen domains in Table \ref{tab:end2end_res}.\footnote{Note that \citet{Zhao2019SimpleQA} reported an accuracy of $85.44\%$. However, they calculate accuracy w.r.t. the correctness of the object entity, which is not standard when testing on the SimpleQuestions dataset (see Section~\ref{sec:eval-metrics}). When we calculate accuracy that way, ~\cite{Petrochuk2018SimpleQuestionsNS} achieves an accuracy of $91.50\%$ and our method achieves $87.31\%$.}

\begin{table}[t]
\centering
\small
\todo{
\caption{Top-1 KGSQA accuracy on seen domains.}
\begin{tabular}{lr}
\toprule
Model                                                                                           & Accuracy (\%) \\ \midrule
Random guess (\cite{bordes2015large})                              & 4.9      \\ \midrule
Memory NN (\cite{bordes2015large})                                 & 62.7     \\ \midrule
Attn. LSTM ( \cite{He2016CharacterLevelQA})                  & 70.9     \\ \midrule
GRU (\cite{lukovnikov2017neural})                             & 71.2     \\ \midrule
BiGRU-CRF \& BiGRU  \cite{Mohammed2018StrongBF} & 74.9     \\ \midrule
CNN \& Attn. CNN \& BiLSTM-CRF \cite{Yin2016SimpleQA}  & 76.4     \\ \midrule
HR-BiLSTM \& CNN \& BiLSTM-CRF \cite{Yu2017ImprovedNR} & 77.0\\ \midrule
\textbf{Ours}                                                                                          & \textbf{77.0}      \\ \midrule
BiLSTM-CRF \& BiLSTM \cite{Petrochuk2018SimpleQuestionsNS}) & 78.1     \\ \midrule
Solr \& TSHCNN  \cite{gupta2018retrieve} & 80.0 \\ \bottomrule
\end{tabular}
\label{tab:end2end_res}
}
    \vspace{-4mm}
\end{table}

\section{Related Work}
\label{sec:related-work}
Methods on the standard KGSQA task are split to those following a pipeline approach--MD, CG, RP \& AS-- (\cite{Tre2016NoNT,Mohammed2018StrongBF,Petrochuk2018SimpleQuestionsNS}) or an end-to-end approach \cite{lukovnikov2017neural,gupta2018retrieve}.
In our work we follow the former approach for solving KGSQA on unseen domains, since we found that all the components except RP are relatively robust for unseen domains.
We leave the exploration of end-to-end approaches for our task for future work.
More related our setting, \citet{Yu2017ImprovedNR} and ~\citet{WuHWZZYC19} tackle RP for KGSQA on unseen relations (instead of whole domains).
Both are model-centric domain adaptation approaches, while ours is data-centric. We experimentally showed that we outperform both in the setting of KGSQA on unseen domains. An interesting direction for future work would be to combine model-centric and data-centric approaches for our task.
More broadly, our work is also related to cross-domain semantic parsing~\cite{su2017cross,yu2018spider,herzig2018decoupling,zhang2019editing}. In contrast to the aforementioned line of work that maps questions to executable logical forms, we focus on questions that can be answered with a single KG fact.

\section{Conclusion}
\label{sec:conclusion}
In this paper, we proposed a data-centric domain adaptation framework for KGSQA that is applicable to unseen domains.
Our framework performs QG to automatically generate synthetic training data for the unseen domains.
We propose a keyword extraction method that when integrated in our QG model, it allows it to generate questions of various lexicalizations for the same underlying relation, thus better resembling the variety of real user questions.
Our experimental results on the SimpleQuestions dataset show that our proposed framework significantly outperforms state-of-the-art zero-shot baselines, and is robust across different domains.
We found that there is room for further improving QG particularly for KGSQA, which is a promising direction for future work.

\section*{Acknowledgements}

This research was supported by
the NWO Innovational Research Incentives Scheme Vidi (016.Vidi.189.039),
the NWO Smart Culture - Big Data / Digital Humanities (314-99-301),
NWO under project nr
CI-14-25, 
the H2020-EU.3.4. - Societal Challenges - Smart, Green And Integrated Transport (814961),
the Google Faculty Research Awards program.
All content represents the opinion of the authors, which is not necessarily shared or endorsed by their respective employers and/or sponsors.

\clearpage

\bibliography{ref}

\begin{thebibliography}{33}
\providecommand{\natexlab}[1]{#1}
\providecommand{\url}[1]{\texttt{#1}}
\expandafter\ifx\csname urlstyle\endcsname\relax
  \providecommand{\doi}[1]{doi: #1}\else
  \providecommand{\doi}{doi: \begingroup \urlstyle{rm}\Url}\fi

\bibitem[Bollacker et~al.(2008)Bollacker, Evans, Paritosh, Sturge, and
  Taylor]{bollacker2008freebase}
Kurt Bollacker, Colin Evans, Praveen Paritosh, Tim Sturge, and Jamie Taylor.
\newblock Freebase: A collaboratively created graph database for structuring
  human knowledge.
\newblock In \emph{SIGMOD}, pages 1247--1250, New York, NY, USA, 2008. ACM.

\bibitem[Bordes et~al.(2015)Bordes, Usunier, Chopra, and
  Weston]{bordes2015large}
Antoine Bordes, Nicolas Usunier, Sumit Chopra, and Jason Weston.
\newblock Large-scale simple question answering with memory networks.
\newblock \emph{CoRR}, abs/1506.02075, 2015.
\newblock URL \url{http://arxiv.org/abs/1506.02075}.

\bibitem[Bota et~al.(2016)Bota, Zhou, and Jose]{bota2016playing}
Horatiu Bota, Ke~Zhou, and Joemon~M Jose.
\newblock Playing your cards right: The effect of entity cards on search
  behaviour and workload.
\newblock In \emph{Proceedings of the 2016 ACM on Conference on Human
  Information Interaction and Retrieval}, pages 131--140. ACM, 2016.

\bibitem[Chu and Wang(2018)]{chu2018survey}
Chenhui Chu and Rui Wang.
\newblock A survey of domain adaptation for neural machine translation.
\newblock In \emph{Proceedings of the 27th International Conference on
  Computational Linguistics, {COLING} 2018, Santa Fe, New Mexico, USA, August
  20-26, 2018}, pages 1304--1319, 2018.
\newblock URL \url{https://www.aclweb.org/anthology/C18-1111/}.

\bibitem[Dong et~al.(2017)Dong, Mallinson, Reddy, and Lapata]{DongMRL17}
Li~Dong, Jonathan Mallinson, Siva Reddy, and Mirella Lapata.
\newblock Learning to paraphrase for question answering.
\newblock In \emph{Proceedings of the 2017 Conference on Empirical Methods in
  Natural Language Processing, {EMNLP} 2017, Copenhagen, Denmark, September
  9-11, 2017}, pages 875--886, 2017.
\newblock URL \url{https://www.aclweb.org/anthology/D17-1091/}.

\bibitem[ElSahar et~al.(2018)ElSahar, Gravier, and Laforest]{elsahar2018zero}
Hady ElSahar, Christophe Gravier, and Fr{\'{e}}d{\'{e}}rique Laforest.
\newblock Zero-shot question generation from knowledge graphs for unseen
  predicates and entity types.
\newblock In \emph{Proceedings of the 2018 Conference of the North American
  Chapter of the Association for Computational Linguistics: Human Language
  Technologies, {NAACL-HLT} 2018, New Orleans, Louisiana, USA, June 1-6, 2018,
  Volume 1 (Long Papers)}, pages 218--228, 2018.
\newblock URL \url{https://www.aclweb.org/anthology/N18-1020/}.

\bibitem[Gupta et~al.(2018)Gupta, Chinnakotla, and
  Shrivastava]{gupta2018retrieve}
Vishal Gupta, Manoj Chinnakotla, and Manish Shrivastava.
\newblock Retrieve and re-rank: A simple and effective ir approach to simple
  question answering over knowledge graphs.
\newblock In \emph{Proceedings of the First Workshop on Fact Extraction and
  VERification (FEVER)}, pages 22--27, 2018.

\bibitem[Han et~al.(2018)Han, Liu, and Sun]{Han0S18}
Xu~Han, Zhiyuan Liu, and Maosong Sun.
\newblock Neural knowledge acquisition via mutual attention between knowledge
  graph and text.
\newblock In \emph{Proceedings of the Thirty-Second {AAAI} Conference on
  Artificial Intelligence, (AAAI-18), the 30th innovative Applications of
  Artificial Intelligence (IAAI-18), and the 8th {AAAI} Symposium on
  Educational Advances in Artificial Intelligence (EAAI-18), New Orleans,
  Louisiana, USA, February 2-7, 2018}, pages 4832--4839, 2018.
\newblock URL
  \url{https://www.aaai.org/ocs/index.php/AAAI/AAAI18/paper/view/16691}.

\bibitem[He et~al.(2016)He, Zhang, Ren, and Sun]{HeZRS16}
Kaiming He, Xiangyu Zhang, Shaoqing Ren, and Jian Sun.
\newblock Deep residual learning for image recognition.
\newblock In \emph{2016 {IEEE} Conference on Computer Vision and Pattern
  Recognition, {CVPR} 2016, Las Vegas, NV, USA, June 27-30, 2016}, pages
  770--778, 2016.
\newblock \doi{10.1109/CVPR.2016.90}.
\newblock URL \url{https://doi.org/10.1109/CVPR.2016.90}.

\bibitem[He and Golub(2016)]{He2016CharacterLevelQA}
Xiaodong He and David Golub.
\newblock Character-level question answering with attention.
\newblock In Jian Su, Xavier Carreras, and Kevin Duh, editors,
  \emph{Proceedings of the 2016 Conference on Empirical Methods in Natural
  Language Processing, {EMNLP} 2016, Austin, Texas, USA, November 1-4, 2016},
  pages 1598--1607. The Association for Computational Linguistics, 2016.
\newblock \doi{10.18653/v1/d16-1166}.
\newblock URL \url{https://doi.org/10.18653/v1/d16-1166}.

\bibitem[Herzig and Berant(2018)]{herzig2018decoupling}
Jonathan Herzig and Jonathan Berant.
\newblock Decoupling structure and lexicon for zero-shot semantic parsing.
\newblock In \emph{Proceedings of the 2018 Conference on Empirical Methods in
  Natural Language Processing}, pages 1619--1629, 2018.

\bibitem[Huang et~al.(2015)Huang, Xu, and Yu]{Huang2015BidirectionalLM}
Zhiheng Huang, Wei Xu, and Kai Yu.
\newblock Bidirectional {LSTM-CRF} models for sequence tagging.
\newblock \emph{CoRR}, abs/1508.01991, 2015.
\newblock URL \url{http://arxiv.org/abs/1508.01991}.

\bibitem[Kingma and Ba(2014)]{kingma2014adam}
Diederik~P Kingma and Jimmy Ba.
\newblock Adam: A method for stochastic optimization.
\newblock \emph{arXiv preprint arXiv:1412.6980}, 2014.

\bibitem[Lample et~al.(2016)Lample, Ballesteros, Subramanian, Kawakami, and
  Dyer]{Lample2016NeuralAF}
Guillaume Lample, Miguel Ballesteros, Sandeep Subramanian, Kazuya Kawakami, and
  Chris Dyer.
\newblock Neural architectures for named entity recognition.
\newblock In \emph{{NAACL} {HLT} 2016, The 2016 Conference of the North
  American Chapter of the Association for Computational Linguistics: Human
  Language Technologies, San Diego California, USA, June 12-17, 2016}, pages
  260--270, 2016.
\newblock URL \url{https://www.aclweb.org/anthology/N16-1030/}.

\bibitem[Lukovnikov et~al.(2017)Lukovnikov, Fischer, Lehmann, and
  Auer]{lukovnikov2017neural}
Denis Lukovnikov, Asja Fischer, Jens Lehmann, and S{\"{o}}ren Auer.
\newblock Neural network-based question answering over knowledge graphs on word
  and character level.
\newblock In \emph{Proceedings of the 26th International Conference on World
  Wide Web, {WWW} 2017, Perth, Australia, April 3-7, 2017}, pages 1211--1220,
  2017.
\newblock \doi{10.1145/3038912.3052675}.
\newblock URL \url{https://doi.org/10.1145/3038912.3052675}.

\bibitem[Mansour et~al.(2009)Mansour, Mohri, and
  Rostamizadeh]{mansour2009domain}
Yishay Mansour, Mehryar Mohri, and Afshin Rostamizadeh.
\newblock Domain adaptation: Learning bounds and algorithms.
\newblock In \emph{{COLT} 2009 - The 22nd Conference on Learning Theory,
  Montreal, Quebec, Canada, June 18-21, 2009}, 2009.
\newblock URL
  \url{http://www.cs.mcgill.ca/\%7Ecolt2009/papers/003.pdf\#page=1}.

\bibitem[Mikolov et~al.(2013)Mikolov, Sutskever, Chen, Corrado, and
  Dean]{mikolov2013distributed}
Tomas Mikolov, Ilya Sutskever, Kai Chen, Gregory~S. Corrado, and Jeffrey Dean.
\newblock Distributed representations of words and phrases and their
  compositionality.
\newblock In \emph{Advances in Neural Information Processing Systems 26: 27th
  Annual Conference on Neural Information Processing Systems 2013. Proceedings
  of a meeting held December 5-8, 2013, Lake Tahoe, Nevada, United States.},
  pages 3111--3119, 2013.
\newblock URL
  \url{http://papers.nips.cc/paper/5021-distributed-representations-of-words-and-phrases-and-their-compositionality}.

\bibitem[Mintz et~al.(2009)Mintz, Bills, Snow, and Jurafsky]{mintz2009distant}
Mike Mintz, Steven Bills, Rion Snow, and Daniel Jurafsky.
\newblock Distant supervision for relation extraction without labeled data.
\newblock In \emph{{ACL} 2009, Proceedings of the 47th Annual Meeting of the
  Association for Computational Linguistics and the 4th International Joint
  Conference on Natural Language Processing of the AFNLP, 2-7 August 2009,
  Singapore}, pages 1003--1011, 2009.
\newblock URL \url{http://www.aclweb.org/anthology/P09-1113}.

\bibitem[Mohammed et~al.(2018)Mohammed, Shi, and Lin]{Mohammed2018StrongBF}
Salman Mohammed, Peng Shi, and Jimmy Lin.
\newblock Strong baselines for simple question answering over knowledge graphs
  with and without neural networks.
\newblock In \emph{Proceedings of the 2018 Conference of the North American
  Chapter of the Association for Computational Linguistics: Human Language
  Technologies, NAACL-HLT, New Orleans, Louisiana, USA, June 1-6, 2018, Volume
  2 (Short Papers)}, pages 291--296, 2018.
\newblock URL \url{https://www.aclweb.org/anthology/N18-2047/}.

\bibitem[Pan and Yang(2010)]{pan2009survey}
Sinno~Jialin Pan and Qiang Yang.
\newblock A survey on transfer learning.
\newblock \emph{{IEEE} Trans. Knowl. Data Eng.}, 22\penalty0 (10):\penalty0
  1345--1359, 2010.
\newblock \doi{10.1109/TKDE.2009.191}.
\newblock URL \url{https://doi.org/10.1109/TKDE.2009.191}.

\bibitem[Pellissier~Tanon et~al.(2016)Pellissier~Tanon, Vrande{\v{c}}i{\'c},
  Schaffert, Steiner, and Pintscher]{pellissier2016freebase}
Thomas Pellissier~Tanon, Denny Vrande{\v{c}}i{\'c}, Sebastian Schaffert, Thomas
  Steiner, and Lydia Pintscher.
\newblock From freebase to wikidata: The great migration.
\newblock In \emph{Proceedings of the 25th international conference on world
  wide web}, pages 1419--1428. International World Wide Web Conferences
  Steering Committee, 2016.

\bibitem[Petrochuk and Zettlemoyer(2018)]{Petrochuk2018SimpleQuestionsNS}
Michael Petrochuk and Luke Zettlemoyer.
\newblock Simplequestions nearly solved: {A} new upperbound and baseline
  approach.
\newblock In \emph{Proceedings of the 2018 Conference on Empirical Methods in
  Natural Language Processing, Brussels, Belgium, October 31 - November 4,
  2018}, pages 554--558, 2018.
\newblock URL \url{https://www.aclweb.org/anthology/D18-1051/}.

\bibitem[Spasojevic et~al.(2017)Spasojevic, Bhargava, and
  Hu]{Spasojevic:2017:DDA:3041021.3053367}
Nemanja Spasojevic, Preeti Bhargava, and Guoning Hu.
\newblock Dawt: Densely annotated wikipedia texts across multiple languages.
\newblock In \emph{Proceedings of the 26th International Conference on World
  Wide Web Companion}, WWW '17 Companion, pages 1655--1662, Republic and Canton
  of Geneva, Switzerland, 2017. International World Wide Web Conferences
  Steering Committee.
\newblock ISBN 978-1-4503-4914-7.
\newblock \doi{10.1145/3041021.3053367}.
\newblock URL \url{https://doi.org/10.1145/3041021.3053367}.

\bibitem[Su and Yan(2017)]{su2017cross}
Yu~Su and Xifeng Yan.
\newblock Cross-domain semantic parsing via paraphrasing.
\newblock In \emph{Proceedings of the 2017 Conference on Empirical Methods in
  Natural Language Processing}, pages 1235--1246, 2017.

\bibitem[T{\"{u}}re and Jojic(2017)]{Tre2016NoNT}
Ferhan T{\"{u}}re and Oliver Jojic.
\newblock No need to pay attention: Simple recurrent neural networks work!
\newblock In \emph{Proceedings of the 2017 Conference on Empirical Methods in
  Natural Language Processing, {EMNLP} 2017, Copenhagen, Denmark, September
  9-11, 2017}, pages 2866--2872, 2017.
\newblock URL \url{https://www.aclweb.org/anthology/D17-1307/}.

\bibitem[Voskarides et~al.(2015)Voskarides, Meij, Tsagkias, de~Rijke, and
  Weerkamp]{voskarides-etal-2015-learning}
Nikos Voskarides, Edgar Meij, Manos Tsagkias, Maarten de~Rijke, and Wouter
  Weerkamp.
\newblock Learning to explain entity relationships in knowledge graphs.
\newblock In \emph{Proceedings of the 53rd Annual Meeting of the Association
  for Computational Linguistics and the 7th International Joint Conference on
  Natural Language Processing (Volume 1: Long Papers)}, pages 564--574,
  Beijing, China, July 2015. Association for Computational Linguistics.
\newblock \doi{10.3115/v1/P15-1055}.
\newblock URL \url{https://www.aclweb.org/anthology/P15-1055}.

\bibitem[Wu et~al.(2019)Wu, Huang, Weng, Zheng, Zhang, Yan, and
  Chen]{WuHWZZYC19}
Peng Wu, Shujian Huang, Rongxiang Weng, Zaixiang Zheng, Jianbing Zhang, Xiaohui
  Yan, and Jiajun Chen.
\newblock Learning representation mapping for relation detection in knowledge
  base question answering.
\newblock In \emph{Proceedings of the 57th Conference of the Association for
  Computational Linguistics, {ACL} 2019, Florence, Italy, July 28- August 2,
  2019, Volume 1: Long Papers}, pages 6130--6139, 2019.
\newblock URL \url{https://www.aclweb.org/anthology/P19-1616/}.

\bibitem[Yih et~al.(2015)Yih, Chang, He, and Gao]{yih2015semantic}
Wen-tau Yih, Ming-Wei Chang, Xiaodong He, and Jianfeng Gao.
\newblock Semantic parsing via staged query graph generation: Question
  answering with knowledge base.
\newblock In \emph{ACL-IJCNLP}, pages 1321--1331. ACL, 2015.

\bibitem[Yin et~al.(2016)Yin, Yu, Xiang, Zhou, and
  Sch{\"{u}}tze]{Yin2016SimpleQA}
Wenpeng Yin, Mo~Yu, Bing Xiang, Bowen Zhou, and Hinrich Sch{\"{u}}tze.
\newblock Simple question answering by attentive convolutional neural network.
\newblock In Nicoletta Calzolari, Yuji Matsumoto, and Rashmi Prasad, editors,
  \emph{{COLING} 2016, 26th International Conference on Computational
  Linguistics, Proceedings of the Conference: Technical Papers, December 11-16,
  2016, Osaka, Japan}, pages 1746--1756. {ACL}, 2016.
\newblock URL \url{https://www.aclweb.org/anthology/C16-1164/}.

\bibitem[Yu et~al.(2017)Yu, Yin, Hasan, dos Santos, Xiang, and
  Zhou]{Yu2017ImprovedNR}
Mo~Yu, Wenpeng Yin, Kazi~Saidul Hasan, C{\'{\i}}cero~Nogueira dos Santos, Bing
  Xiang, and Bowen Zhou.
\newblock Improved neural relation detection for knowledge base question
  answering.
\newblock In \emph{Proceedings of the 55th Annual Meeting of the Association
  for Computational Linguistics, {ACL} 2017, Vancouver, Canada, July 30 -
  August 4, Volume 1: Long Papers}, pages 571--581, 2017.
\newblock \doi{10.18653/v1/P17-1053}.
\newblock URL \url{https://doi.org/10.18653/v1/P17-1053}.

\bibitem[Yu et~al.(2018)Yu, Zhang, Yang, Yasunaga, Wang, Li, Ma, Li, Yao,
  Roman, et~al.]{yu2018spider}
Tao Yu, Rui Zhang, Kai Yang, Michihiro Yasunaga, Dongxu Wang, Zifan Li, James
  Ma, Irene Li, Qingning Yao, Shanelle Roman, et~al.
\newblock Spider: A large-scale human-labeled dataset for complex and
  cross-domain semantic parsing and text-to-sql task.
\newblock In \emph{Proceedings of the 2018 Conference on Empirical Methods in
  Natural Language Processing}, pages 3911--3921, 2018.

\bibitem[Zhang et~al.(2019)Zhang, Yu, Er, Shim, Xue, Lin, Shi, Xiong, Socher,
  and Radev]{zhang2019editing}
Rui Zhang, Tao Yu, Heyang Er, Sungrok Shim, Eric Xue, Xi~Victoria Lin, Tianze
  Shi, Caiming Xiong, Richard Socher, and Dragomir Radev.
\newblock Editing-based sql query generation for cross-domain context-dependent
  questions.
\newblock In \emph{Proceedings of the 2019 Conference on Empirical Methods in
  Natural Language Processing and the 9th International Joint Conference on
  Natural Language Processing (EMNLP-IJCNLP)}, pages 5341--5352, 2019.

\bibitem[Zhao et~al.(2019)Zhao, Chung, Goyal, and Metallinou]{Zhao2019SimpleQA}
Wenbo Zhao, Tagyoung Chung, Anuj~Kumar Goyal, and Angeliki Metallinou.
\newblock Simple question answering with subgraph ranking and joint-scoring.
\newblock In \emph{Proceedings of the 2019 Conference of the North American
  Chapter of the Association for Computational Linguistics: Human Language
  Technologies, {NAACL-HLT} 2019, Minneapolis, MN, USA, June 2-7, 2019, Volume
  1 (Long and Short Papers)}, pages 324--334, 2019.
\newblock URL \url{https://aclweb.org/anthology/papers/N/N19/N19-1029/}.

\end{thebibliography}
\bibliographystyle{plainnat}
\end{document}